\documentclass[a4paper, 10 pt, conference]{ieeeconf}  % Comment this line out
\IEEEoverridecommandlockouts                              % This command is only
\usepackage[utf8]{inputenc}
\usepackage[T1]{fontenc}

\usepackage{amsmath}
\usepackage{cite}
\usepackage{comment}
\usepackage{subcaption}
\usepackage{ctable}
\usepackage{breakurl}
\usepackage{amsfonts}
\usepackage{amssymb}
\usepackage{url}

\makeatletter
\DeclareMathOperator{\is}{IS}
\DeclareMathOperator*{\mean}{\mathbb{E}}

\newcommand{\AL}[1]{{\color{black}#1}}
\title{\LARGE \bf
The Six Fronts of the Generative Adversarial Networks
}

\author{Alceu Bissoto, Eduardo Valle, and Sandra Avila$^\ast$ \\
RECOD Lab., University of Campinas (UNICAMP), Brazil\\
\thanks{
        A. Bissoto and S. Avila are with the Institute of Computing (IC/UNICAMP), and E. Valle is with the School of Electrical and Computing Engineering (FEEC/UNICAMP).
        All authors are affiliated to the RECOD Lab.
        $^\ast$Contact author: sandra@ic.unicamp.br.
    }
}

\begin{document}

\maketitle
\thispagestyle{empty}
\pagestyle{empty}

%%%%%%%%%%%%%%%%%%%%%%%%%%%%%%%%%%%%%%%%%%%%%%%%%%%%%%%%%%%%%%%%%%%%%%%%%%%%%%%%
\begin{abstract}
Generative Adversarial Networks fostered a newfound interest in generative models, resulting in a swelling wave of new works that new-coming researchers may find formidable to surf. In this paper, we intend to help those researchers, by splitting that incoming wave into six ``fronts'': Architectural Contributions, Conditional Techniques, Normalization and Constraint Contributions, Loss Functions, Image-to-image Translations, and Validation Metrics. The division in fronts organizes literature into approachable blocks, ultimately communicating to the reader how the area is evolving. Previous surveys in the area, which this works also tabulates, focus on a few of those fronts, leaving a gap that we propose to fill with a more integrated, comprehensive overview.  
Here, instead of an exhaustive survey, we opt for a straightforward review: our target is to be an entry point to this vast literature, and also to be able to update experienced researchers to the newest techniques.
\end{abstract}
%%%%%%%%%%%%%%%%%%%%%%%%%%%%%%%%%%%%%%%%%%%%%%%%%%%%%%%%%%%%%%%%%%%%%%%%%%%%%%%%

\section{Introduction}
Generative Adversarial Networks (GANs) is a hot topic in machine learning. Its flexibility enabled GANs to be used to solve different problems, ranging from generative tasks, such as image synthesis~\cite{karras2019style, brock2018large, bissoto2018skin}, style-transfer~\cite{park2019semantic}, super-resolution~\cite{ledig2017photo, wang2018esrgan}, and image completion~\cite{yu2018generative}, to decision tasks, such as classification~\cite{salimans2016} and segmentation~\cite{xue2018segan}. Also, the ability to reconstruct objects and scenes challenge the ability of computer vision solutions to visually represent what they are trying to learn.

The success of the method proposed by Goodfellow \emph{et al.}~\cite{goodfellow2014} dragged attention from academia, which favored GANs over other generative models. The most popular generative models apart from GANs are Variational Autoencoders~\cite{kingma2013auto} and Autoregressive models~\cite{oord2016pixel,van2016conditional}. All three are based on the principle of maximum likelihood. Both autoregressive models and variational autoencoders work to find explicit density models. This makes it challenging to capture the complexity of a whole domain while keeping the model feasible to train and evaluate. To overcome the feasibility problems, autoregressive models decompose the $n$-dimensional probability distribution into a product of one-dimensional probability distribution, while variational autoencoders use approximations to the density function. Differently from these methods, GANs use implicitly density functions, which are modeled in the networks that compose its framework. 

Since autoregressive models decompose the probability distribution, every new sample needs to be generated one entry at a time. So for an image, every new pixel takes into consideration the previously generated ones. This process makes the generation process slow, and it can not be parallelized. Variational autoencoders and GANs do not suffer from the same problem, but variational autoencoders' generated samples are regarded as producing lower-quality samples (even though measuring quality of synthetic samples is often subjective).
Despite each method have its strengths and flaws, the fast evolution of GAN methods enabled GANs to surpass other methods in most of their strengths, and today it is the most studied generative model, with the number of papers growing each year rapidly.

According to Google Scholar, to this date, the seminal paper from Goodfellow \emph{et al.} (``Generative Adversarial Nets'') was cited more than $12,000$ times, and since 2017 the pace accelerated significantly. For this reason, surveys that can make sense of the evolution of related works are necessary. \AL{In 2016, Goodfellow \cite{goodfellow2016tutorial} in his NeurIPS tutorial, characterized different generative models, explained challenges that are still relevant today, and showcased different applications that could benefit by using GANs. In 2017, Cresswell \textit{et al.} \cite{creswell2018generative} summarized different techniques, especially for the signal processing community, highlighting techniques that were still emerging, and that are a lot more mature today. Wang \textit{et al.} \cite{wang2019generative} recently reviewed architectural, and especially, loss advancements.}
However, a \AL{multifaceted} review of different trends of thought that guided authors is lacking. We attempt to fill this gap with this work. We do not intend to provide an exhaustive and extensive literature review, but instead, to group relevant works that influenced the literature, highlighting their contribution to the overall scene. 

We divide the advancements in six fronts --- Architectural (Section~\ref{sec:sotaarch}), Conditional Techniques (Section~\ref{sec:sotaclass}), Normalization and Constraint (Section~\ref{sec:sotanorm}), Loss Functions (Section~\ref{sec:sotaloss}), Image-to-image Translation (Section~\ref{sec:sotatranslate}), and Validation (Section~\ref{sec:sotavalidation}) --- providing a comprehensive notion of how the scenario evolved, showing trends that resulted to GANs being capable of generating images indistinguishable from real photos.

\begin{figure*}[t]
    \centering
    \includegraphics[width=\linewidth, trim={0.2cm 0 0.2cm 0}, clip,]{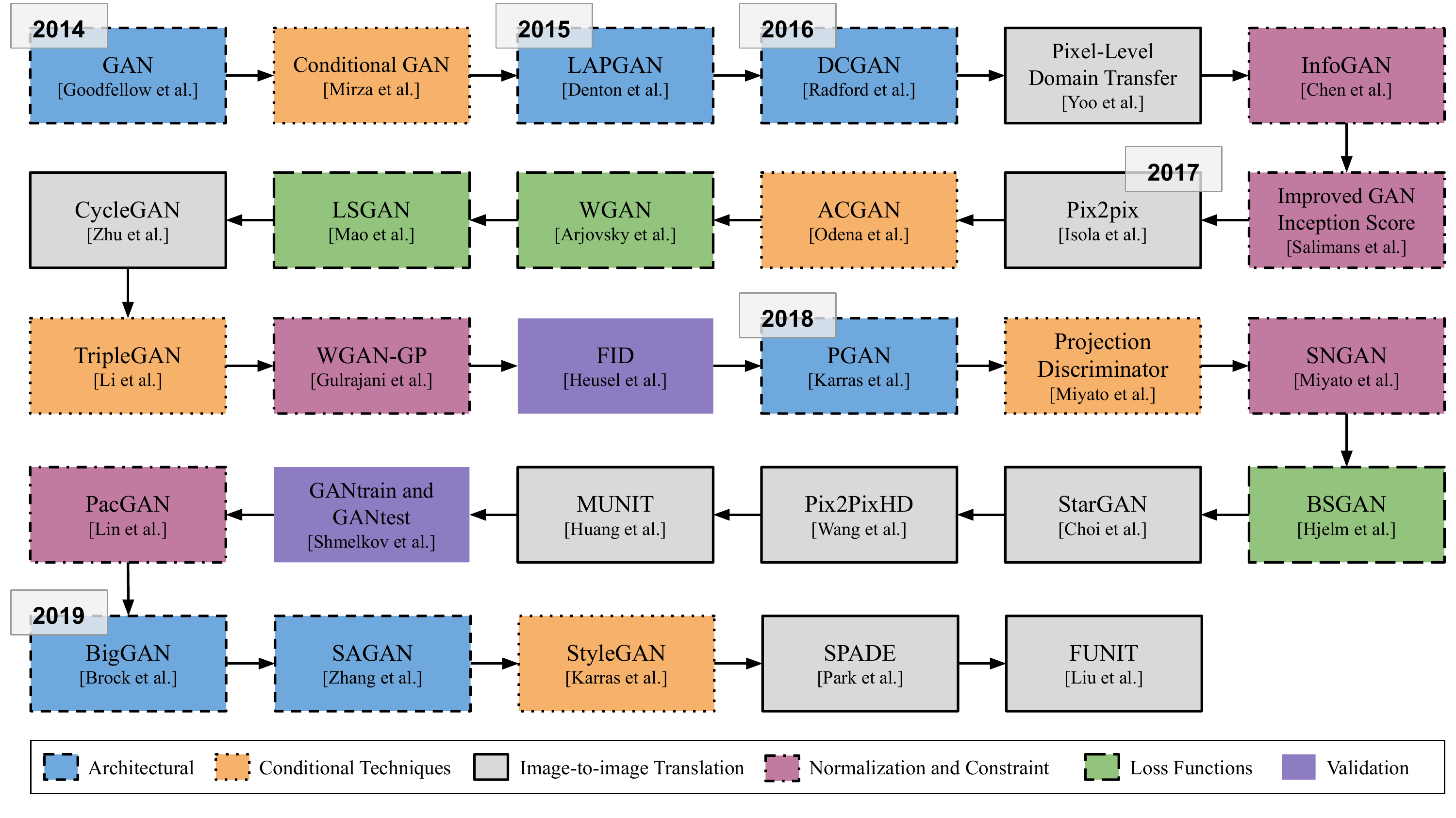}
    \caption{Timeline of the GANs covered in this paper. Just like our text, we split it in six fronts (architectural, conditional techniques, normalization and constraint, loss functions, image-to-image translation and validation metrics), each represented by a different color and a different line/border style. 
    }
    \label{fig:timeline}
\end{figure*}

Since we choose an evolutionary view of literature, we lose most of its chronology. To recover the time dimension, we chronologically organize the GANs in Figure \ref{fig:timeline} while also highlighting by their main contribution, linking them to one of the six fronts of the GANs.

\section{Basic Concepts}

Before introducing the formal concept of GANs, we start with an intuitive analogy proposed by Dietz \cite{dietz}. The scenario is a boxing match, with a coach and two boxers. Let us call the fighters \textbf{Gabriel} and \textbf{Daniel}. Both boxers learn from each other during the match, but only \textbf{Daniel} has a coach. \textbf{Gabriel} only learns from \textbf{Daniel}. Therefore, at the beginning of the boxing, \textbf{Gabriel} keeps himself focused, observing his adversary and his moves, trying to adapt every round, guessing the teachings the coach gave to \textbf{Daniel}. After many rounds, \textbf{Gabriel} was able to learn the fundamentals of boxing during the match against \textbf{Daniel} and ideally, the boxing match would have 50/50 odds.

In the boxing analogy, \textbf{Gabriel} is the generator ($G$), \textbf{Daniel} is the discriminator ($D$), and the coach is the real data --- larger the data, more experienced the coach. Goodfellow \emph{et al.} \cite{goodfellow2014} introduced GANs as two deep neural networks (the generator $G$ and the discriminator $D$) that play a minimax two-player game with value function $V(D, G)$ as follows:

\begin{equation}
\begin{split}
\min_{G} \max_{D}V(D,G) = & \mathbb{E}_{x \sim p_{data}(x)}[\log{D(x)}] + \\
& \mathbb{E}_{z \sim p_{z}(z)}[\log{(1 - D(G(z)))}],    
\end{split}
\label{eq:originalloss}
\end{equation}

\noindent where $z$ is the noise, $p_z$ is the noise distribution, $x$ is the real data, and $p_{data}$ is the real data distribution.  

The goal of the generator is to create samples as they were from the real data distribution $p_{data}$. To accomplish that, it learns the distribution of the real data and applies the learned mathematical function to a given noise $z$ from the distribution $p_z$. The goal of the discriminator is to be able to discriminate between real (from the real data distribution) and generated samples (from the generator) with high precision.

\begin{figure}[h]
    \centering
    \includegraphics[trim={1.2cm 0 0.25cm 0}, clip, width=\linewidth]{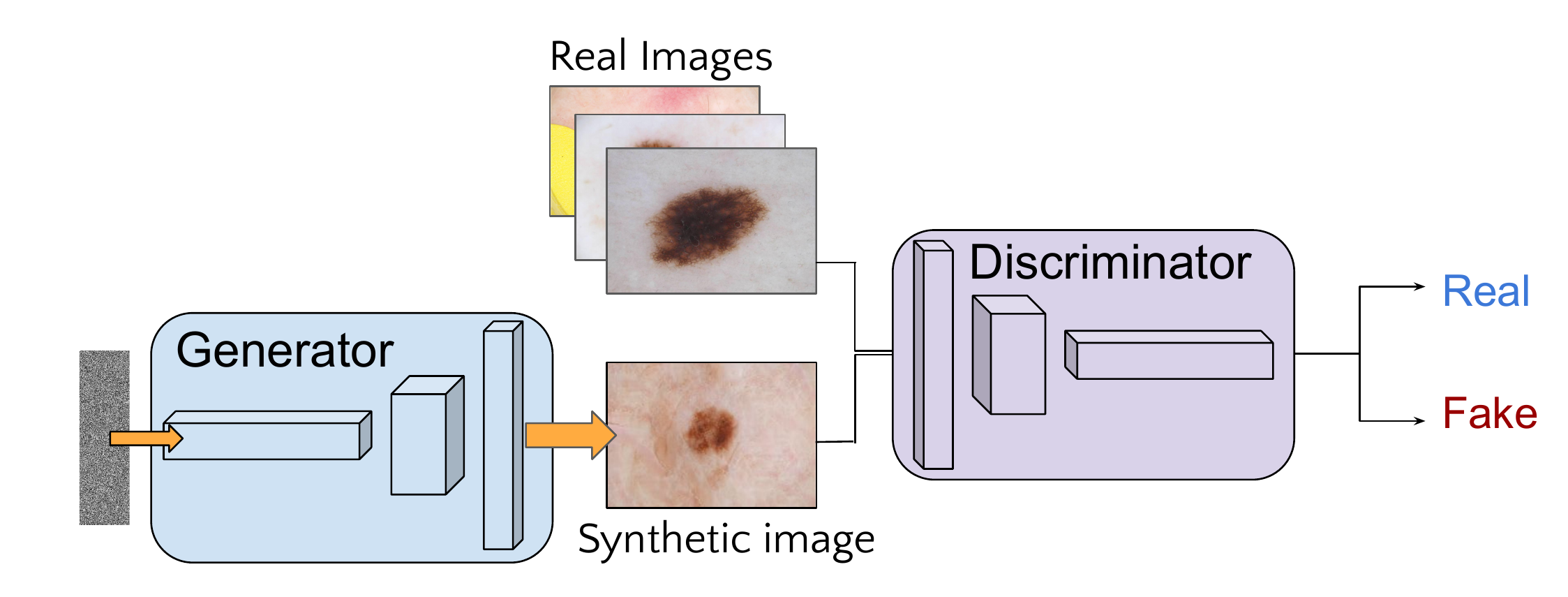}
    \caption{Simplified GAN Architecture. GANs are composed of two networks that are trained in a competition. While the generator learns to transform the input noise into samples that could belong to the target dataset, the discriminator learns to classify the images as real or fake. Ideally, after enough training, the discriminator should not be able to differentiate between real and fake images, since the generator would be synthesizing good quality images with high variability.}
    \label{fig:arch_vanilla}
\end{figure}

During training, the generator receives feedback from the discriminator's decision (that classify the generated sample as real or fake), learning how to fool the discriminator better the next time. In Figure \ref{fig:arch_vanilla}, we show a simplified GAN pipeline.

\section{GANs' Challenges}

In 2017, GANs suffered from high instability and were considered hard to train~\cite{arjovsky2017}. Since then, different architectures, loss functions, conditional techniques, and constrain methods were introduced, easing the convergence of GAN models. However, there are still hyperparameter choices that may highly influence training. The batch size and layers width were recently in the spotlight after BigGAN \cite{brock2018large} showed state-of-the-art results for ImageNet image synthesis by increasing radically these, among other factors. Aside from its impressive results, it showed directions to improve the GAN framework further. 

Is the gradient noise introduced when using small mini-batches more impactful than the problems caused by the competition between discriminator and generator? How much can GAN benefit (if it can at all) from scaling to very deep architectures and parallelism, and vast volumes of data? It seems that the computational budget can be critical for the future of GANs. Lucic \emph{et al.}~\cite{lucic2018gans} showed that given enough time for hyperparameter tuning and random restarts, despite multiple proposed losses and techniques, different GANs can achieve the same performance.

However, even in normal conditions, GANs achieve incredible performance for specific tasks such as face generation, but the same quality is not perceived when dealing with more general datasets such as ImageNet. The characteristics of the dataset that favor GANs performance still uncertain. A possibility is the regularity (using the same object poses, placement on the image, or using different objects with the same characteristics) are easier than others where variation is high such ImageNet.

A factor that is related to data regularity and variability is the number of classes. GANs suffer to cover all class possibilities of a target dataset if it is unbalanced (\emph{e.g.}, medical datasets) or if the class count is too high (\emph{e.g.}, ImageNet). This phenomenon is called ``mode collapse'' \cite{arjovsky2017}, and despite efforts from the literature to mitigate the issue \cite{salimans2016,lin2018pacgan}, modern GAN solutions still display this undesired behavior. If we think in the use case where we want to augment a training dataset with synthetic images,
it is even more impactful once the object classes we want to generate are usually the most unbalanced ones.

Validating synthetic images is also challenging. Qualitative evaluation, where grids of low-resolution images are compared side-by-side, was (and still is) one of the most used methods for performance comparison between different works. Authors also resort to services like the Amazon Mechanical Turk (AMT), which enable to perform statistical analysis on multiple human annotators' choices over synthetic samples. However, except for the cases where the difference is massive, the subjective nature of this approach may lead to wrong decisions. 

Ideally, we want quantitative metrics that consider different aspects of the synthetic images, such as the overall structure of the object, the presence of fine details, and variability between samples. The most accepted metrics, Inception Score (IS) \cite{salimans2016} and Frech\`et Inception Distance (FID) \cite{heusel2017gans}, both rely on the activations of an ImageNet pre-trained Inceptionv3 network to output its scores. This design causes scores to be unreliable, especially for contexts that are not similar to ImageNet's. 

Other metrics such as GANtrain and GANtest can analyze those mentioned aspects, however we have to consider the possible flaws of the used classification networks --- especially bias --- which could reinforce its presence in the synthetic images. Thus, only by employing diversified metrics, evaluation methods, and content-specific measures, we can truly assess the quality of the synthetic samples.

\section{GAN Literature Review}
\subsection{Architectural Methods} \label{sec:sotaarch}

Following the seminal paper of Goodfellow \emph{et al.}, many works proposed \textit{architectural enhancements}, which enabled exploring GANs
in different contexts. At that time, GANs were capable only of generating low resolution samples ($32 \times 32$) from simpler datasets like MNIST \cite{lecun1998mnist} and Faces \cite{susskind2010toronto}. However, in 2016, crucial architectural changes were proposed, boosting GANs research and increasing the complexity and quality of the synthetic samples. 

\begin{figure}[b]
    \centering
    \includegraphics[width=\linewidth]{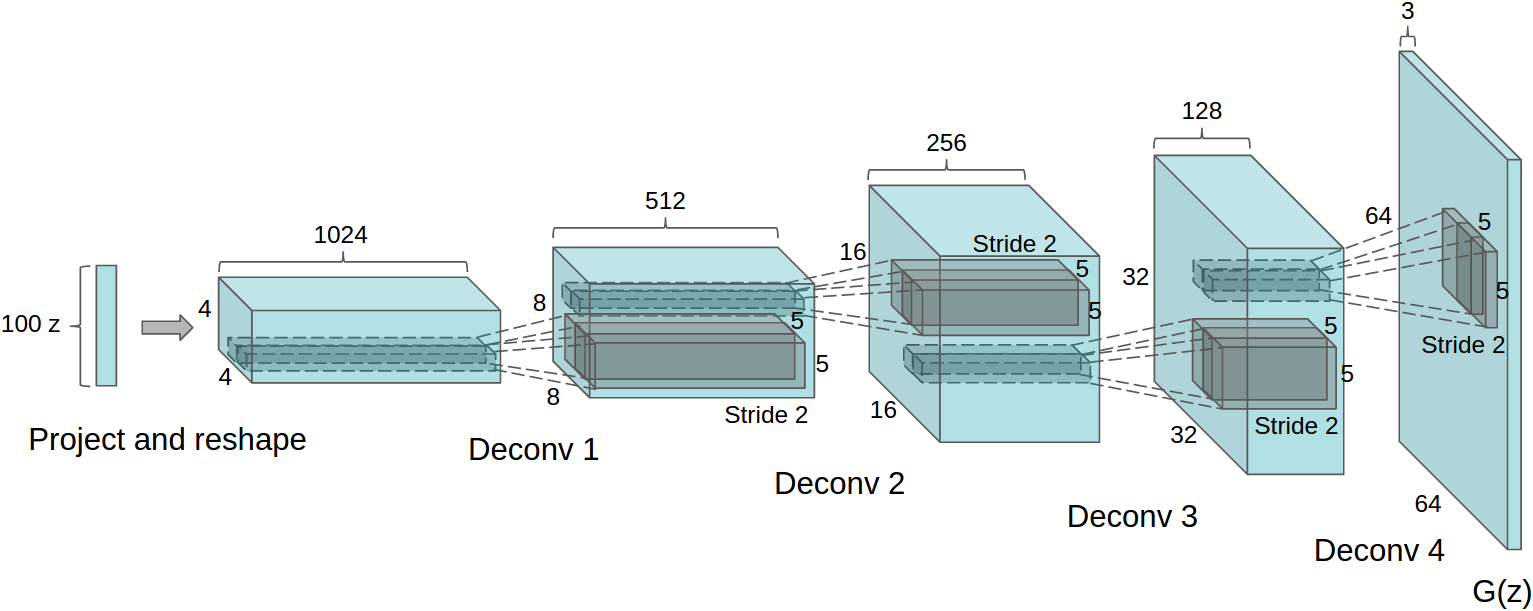}
    \caption{DCGAN's generator architecture, reproduced from Radford \emph{et al.} \cite{radford2015dcgan}. The replacement of fully-connected and pooling layers by (strided and fractional-strided) convolutional layers enabled GANs architecture to be deeper and more complex. }
    \label{fig:arch_dcgan}
\end{figure}

Deep Convolutional GAN (DCGAN) \cite{radford2015dcgan} proposed detailed architectural guidelines that stabilized GAN's training, enabling the use of deeper models and achieving higher resolutions (see Figure \ref{fig:arch_dcgan}). The proposals to remove pooling and fully-connected layers guided the future models' design; while the proposal of using batch normalization inspired other normalization techniques \cite{miyato2018spectral, karras2017progressive} and still is used in modern GAN frameworks \cite{miyato2018cgans}.
DCGAN caused such an impact in the community that it is still used nowadays when working with simple low-resolution datasets and as an entry point when applying GANs in new contexts.

At the same period, Denton \emph{et al.} \cite{denton2015deep} proposed the Laplacian Pyramid GAN (LAPGAN): an incremental architecture where the resolution of the synthetic sample is progressively increased throughout the generation pipeline. This modification enabled the generation of synthetic images of a resolution of up to $96\times96$ pixels. 

\begin{figure}[t]
    \centering
    \includegraphics[width=\linewidth]{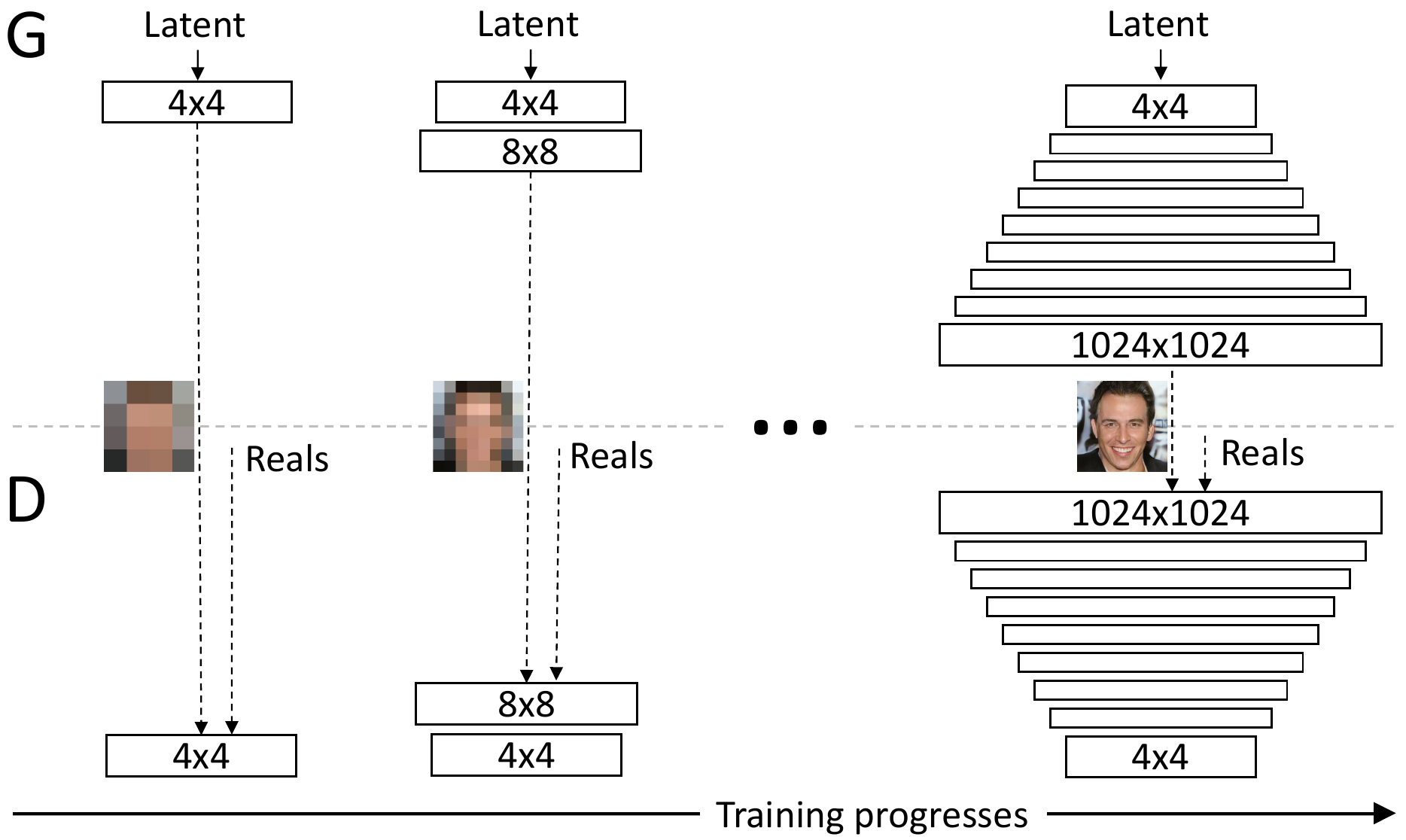}
    \caption{Simplified PGAN's Incremental Architecture. PGAN starts by generating and discriminating low-resolution samples, adjusting the generation coarse details. Progressively, the resolution is increased, enhancing the synthetic image fine details. Reproduced from Karras \emph{et al.} \cite{karras2017progressive}.\\}
    \label{fig:arch_pgan}
\end{figure}

In 2018, this type of architecture gained popularity, and it is still employed for improved stability and high-resolution generation. Progressive GAN (PGAN) \cite{karras2017progressive} improved the incremental architecture to generate human faces of $1024\times1024$ pixels. While the spatial resolution of the generated samples increases, layers are progressively added to both Generator and Discriminator (see Figure \ref{fig:arch_pgan}). Since older layers remain trainable, generation happens at different resolutions for the same image. It enables coarse/structural image details to be adjusted in lower resolution layers, and fine details in higher resolution layers. 

\begin{figure}[t]
    \centering
    \includegraphics[width=\linewidth]{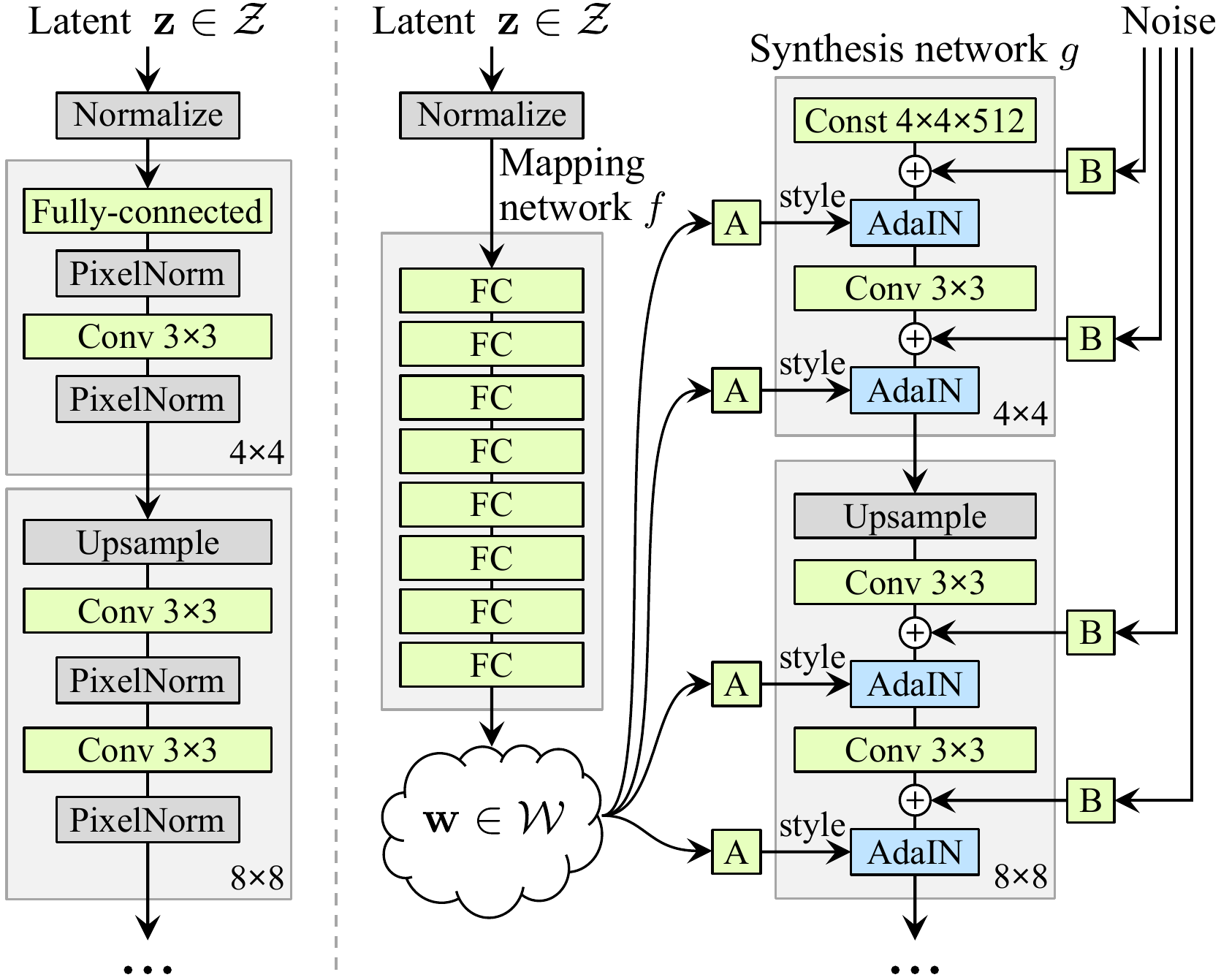}
    \caption{Comparison between generators on PGAN \cite{karras2017progressive} (on the left) and StyleGAN (on the right). First, a sequence of fully-connected layers is used to extract style information and disentangle the noise and class information. Next, style information feeds AdaIN layers, providing its parameters for shift and scale. Also, different sources of noise feed each layer to simplify the task of controlling stochastic aspects of generation (such as placement of hair and skin pores). Reproduced from Karras \emph{et al.} \cite{karras2019style}.}
    \label{fig:arch_stylegan}
\end{figure}

Very recently, StyleGAN \cite{karras2019style} showed remarkable performance when generating human faces, dethroning PGAN in this task. Despite keeping the progressive training procedure, several changes were proposed (see Figure \ref{fig:arch_stylegan}).
The authors changed how information is usually fed into the generators: In other works, the main input of noise passes through the whole network, being transformed layer after layer until becoming the final synthetic sample. In StyleGAN, the generator receives information directly in all its layers. Before feeding the generator, the input label data goes through a mapping network (composed of several sequential fully-connected layers) that extract class information. This information, called ``style'' by the authors, is then combined with the input latent vector (noise). Then, it is incorporated in all the generator's layers by providing the Adaptive Instance Normalization (AdaIN) \cite{huang2017arbitrary} layer's parameters for scale and~bias.

Also, the authors added independent sources of noise that feed different layers of the generator. By doing this, the authors expect that each source of noise can control a different stochastic aspect of the generation (\emph{e.g.}, placement of hair, skin pores, and background) enabling more variation and higher detail level.

An effect of these modifications, specially of the mapping network, is the disentanglement of the relations between the noise and the style. This enables subtle modifications on the noise to result in subtle changes to the generated sample while keeping it plausible, improving the quality of the generated images, and stabilizing training. 

The same idea of disentangling the class and latent vectors was first explored by InfoGAN \cite{chen2016infogan}, where the network was responsible for finding characteristics that could control generation and disentangle this information from the latent vector. This disentanglement allows discovered (by the network itself) features, which are embedded in the latent vector, to control visual aspects of the image, making the generation process more coherent and tractable.

Another class of architectural modifications consists of GANs that were expanded to include different networks in the framework.
Encoders are the most common component aside from the original generator and discriminator. Their addition to the framework created an entire new line of possibilities doing image-to-image translation \cite{pix2pix} (detailed later in Section \ref{sec:sotatranslate}). Also, super-resolution and image segmentation solutions use encoders in their architecture. Since encoders are present in other generative methods such as Variational Autoencoders (VAE) \cite{kingma2013auto}, there were also attempts of extracting the best of both generative methods~\cite{larsen2016autoencoding} by combining them.
Finally, researchers also included classifiers to be trained jointly with the generator and discriminator, improving generation~\cite{nguyen2017plug}, and semi-supervised capabilities \cite{chongxuan2017triple}.

\subsection{Conditional Techniques} \label{sec:sotaclass}

In 2014, Mirza \emph{et al.} \cite{mirza2014} introduced a way to control the class of the generated samples: concatenate the label information to the generator's input noise. However simple, this method instigated researchers to be creative with GAN inputs, and enabled GANs to solve different and more complex tasks.

The next step was to include the label in the loss function. Salimans \emph{et al.} \cite{salimans2016} proposed to increase the output size of the discriminator, to include classes, for semi-supervised classification purposes. Despite having a different objective, it inspired future works. Odena \emph{et al.} \cite{odena2017conditional} focused on generation with Auxiliary Classifier GAN (ACGAN), splitting both tasks (label and real/fake classification) for the discriminator, while still feeding the generator with class information. It greatly improved generation, creating $128\times128$ synthetic samples (surpassing LAPGAN's $96\times96$ resolution) for all $1000$ classes of ImageNet. Also, authors showed that GANs benefit from higher resolution generation, which increases discriminability for a classification network. 

The next changes again modified how the class information is fed to the network. On TripleGAN \cite{chongxuan2017triple} the class information is concatenated in all layers' feature vectors, except for the last one, on both discriminator and generator. Note that sometimes the presented approaches can be combined: concatenation became the standard method of feeding conditional information to the network, while using ACGAN's loss function component. 

Next, Miyato \emph{et al.} \cite{miyato2018cgans} introduced a new method to feed the information to the discriminator. The class information is first embedded, and then integrated to the model through an inner product operation at the end of the discriminator. In their solution, the generator continued concatenating the label information to the layers' feature vectors. Although the authors show the exact networks used in their experiments, this technique is not restricted and can be applied to any GAN architecture for different tasks.

Recently, an idea used for style transfer was incorporated to the GAN framework.
In style transfer the objective is to transform the source image in a way it resembles the style (especially the texture) of a given target image, without losing its content components. The content comprehends information that can be shared among samples from different domains, while style is unique to each domain. When considering paintings for example, the ``content'' represent the objects in the scene, with its edges and dimensions; while ``style'' can be the artists' painting (brushing) technique and colors used.    
Adaptive Instance Normalization (AdaIN) \cite{huang2017arbitrary} is used to infuse a style (class) into content information. The idea is to \textit{first} normalize the feature maps according to its dimensions' mean and variance evaluated for each sample, and each channel. Intuitively, Huang \emph{et al.} explained that at this point (which is called simply Instance Normalization), the style itself is normalized, being appropriate to receive a new style. Finally, AdaIN uses statistics obtained from a style encoder to scale and shift the normalized content, infusing the target style into the instance normalized features. 

This idea was adapted and enhanced for current state-of-the-art methods for plain generation \cite{huang2018multimodal, karras2019style}, image-to-image translation \cite{park2019semantic}, and to few-shot image-translation \cite{liu2019few}. Authors usually employ an encoder to extract style that feeds a multi-layer perceptron (MLP) that outputs the statistics used to control scale and shift on AdaIN layers.  

In Spatially-Adaptive Normalization (SPADE) \cite{park2019semantic}, a generalization of previous normalizations (\emph{e.g.}, Batch Normalization, Instance Normalization, Adaptive Instance Normalization) is used to incorporate the class information into the generation process, but it focuses at working with input semantic maps for image-to-image translation. The parameters used to shift and scale the feature maps are tensors that preserve spatial information from the input semantic map. Those parameters are obtained through convolutions, then multiplied and summed element-wise to the feature map. Since this information is included in most of the generator's layers, it prevents the map information to fade away during the generation process.

\subsection{Normalization and Constraint Techniques} \label{sec:sotanorm}

On DCGAN \cite{radford2015dcgan}, the authors advocated the use of batch normalization \cite{ioffe2015batch} layers on both generator and discriminator to reduce internal covariate shift. It was the beginning of the use of normalization techniques, which also developed with time, for increased stability. In batch normalization, the output of a previous activation layer is normalized using the current batch statistics (mean and standard variation). Then, it adds two trainable parameters to scale and shift the normalization result.

For PGAN \cite{karras2017progressive} authors, the problem is not the internal covariate shift, but the signal magnitudes that explode due to competition between the generator and discriminator. To solve this problem, they moved away from batch normalization, and introduced two techniques to constrain the weights. On \textit{Pixelwise Normalization} (see Equation \ref{eq:pixelnorm}), they normalize the feature vector in each pixel. This approach is used in the generator, and do not add any trainable parameter to the network. 

\begin{equation}
    b_{x,y} = a_{x,y} / \sqrt{\frac{1}{N} \sum_{j=0}^{N-1} (a^j_{x,y})^2 + \epsilon},
    \label{eq:pixelnorm}
\end{equation}
\noindent where $\epsilon=10^{-8}$, $N$ is the number of feature maps, and $a_{x,y}$ and $b_{x,y}$ are the original and normalized feature vector in pixel $(x,y)$, respectively.

The other technique introduced by Karras \emph{et al.} \cite{karras2017progressive} is called \textit{Equalized Learning Rate} (see Equation \ref{eq:eqlr}).

\begin{equation}
    \hat{w}_i = w_i / c,    
    \label{eq:eqlr}
\end{equation}
\noindent where $w_i$ are the weights and $c$ is the per-layer normalization constant from He's initializer~\cite{he2015delving}. During initialization, weights are all sampled from the same distribution $\mathcal{N}(0,1)$ at runtime, and then scaled by $c$. It is used to avoid weights to have different dynamic ranges across different layers. This way, the learning rate impacts all the layers by the same factor, avoiding it to be too large for some, and too little for~others. 

\textit{Spectral Normalization} \cite{miyato2018spectral} also acts constraining weights, but only at the discriminator. The idea is to constrain the Lipschitz constant of the discriminator by restricting the spectral norm of each layer. By constraining its Lipschitz constant, it limits how fast the weights of the discriminator can change, stabilizing training. For this, every weight~$W$ is scaled by the largest singular value of $W$. This technique is currently present in state-of-the-art networks \cite{park2019semantic, zhang2018self}, being applied on both generator and discriminator.

So far, the presented constraint methods concern about normalizing weights. Differently, \emph{Gradient penalty} \cite{gulrajani2017improved} enforces 1-Lipschitz constraint to make all gradients with norm at most $1$ everywhere. It adds an extra term to the loss function to penalize the model if the gradient norm goes beyond the target value $1$.

Some methods did not change loss functions or added normalization layers to the model. Instead, those methods introduced subtle changes in the training process to deal with GAN's general problems, such as training instability and low variability on the generated samples. 
\textit{Minibatch Discrimination} \cite{salimans2016} gives the discriminator information around the mini-batch that is being analyzed. Roughly speaking, this is achieved by attaching a component\footnote{Vector of differences between the sample being analyzed and the others present in the mini-batch.} to an inner layer of the discriminator. With this information, the discriminator can compare the images on the mini-batch, forcing the generator to create images that are different from each other. 

Similarly with respect of giving more information to the discriminator, PacGAN \cite{lin2018pacgan} packs (concatenates in the width axis) different images from the same source (real or synthetic) before feeding the discriminator. According to the authors, this procedure helps the generator to cover all target labels in the training data, instead of limiting itself to generate samples of a single class that are able to fool the discriminator (a problem called \textit{mode collapse} \cite{arjovsky2017}).

\subsection{Loss Functions} \label{sec:sotaloss}

Theoretical advances towards understanding GAN's training and the sources of its instabilities \cite{arjovsky2017} pointed that the Jensen-Shannon Divergence (JSD) (which is used in GAN's formulation to measure similarity between the real data distribution and the generator's) is responsible for vanishing gradients when the discriminator is already well trained. This theoretical understanding contributed to motivate next wave of works, that explored alternatives to the JSD.

Instead of JSD, authors proposed using the Pearson $\chi^2$ (Least Squares GAN)\cite{mao2016}, the Earth-Mover Distance (Wasserstein GAN) \cite{arjovsky2017wasserstein}, and Cramér Distance (Cramér GAN) \cite{bellemare2017cramer}. 
One core principle explored was to penalize samples even when it is on the correct side of the decision boundary, avoiding the vanishing gradients problem during training.

Other introduced methods choose to keep the divergence function intact and introduce components to the loss function to increase image quality, training stability, or to deal with mode collapse and vanishing gradients. Those methods often can be employed together (and with different divergence functions) evidencing the many possibilities of tuning GANs when working on different contexts.

An example that shows the possibility of joining different techniques is the Boundary-Seeking GAN (BSGAN) \cite{hjelm2017}, where a simple component (which must be adjusted for different f-divergence functions) tries to guide the generator to generate samples that make the discriminator output $0.5$ for every sample.

\textit{Feature Matching} \cite{salimans2016} comprises a new loss function component for the generator that induces it to match the features that better describe the real data. Naturally, by training the discriminator we are asking it to find these features, which are present in its intermediate layers.
Similarly to Feature Matching, \textit{Perceptual Loss} \cite{johnson2016perceptual} also uses statistics from a neural network to compare real and synthetic samples and encourage them to match. Differently though, it employs ImageNet pre-trained networks (VGG \cite{simonyan15} is often used), and add an extra term to the loss function. This technique is commonly used for super-resolution methods, and also image-to-image translation \cite{wang2017high}.

Despite all the differences between the distinct loss function, and methods used to train the networks, given enough computational budget, all can reach comparable performance~\cite{lucic2018gans}. However, since solutions are more urgent than ever, and GANs have the potential to impact multiple areas, from data augmentation, image-to-image translation, super-resolution and many others, gathering the correct methods that enable a fast problem-solver solution is essential.

\subsection{Image-to-image Translation Methods} \label{sec:sotatranslate}

The addition of encoders in the architecture enabled GANs for image-to-image translation starting with Yoo \emph{et al.} \cite{yoo2016pixel}, in 2016. The addition of the encoder to the generator network transformed it into an encoder-decoder network (autoencoder). Now, the source image is first encoded into a latent representation, which is then mapped to the target domain by the generator.
The changes in the discriminator are not structural, but the task changed. In addition to the traditional adversarial discriminator, the authors introduce a \textit{domain discriminator} that analyze pairs of source and target (real and fake) samples and judge if they are associated or~not.

\begin{figure}[t]
    \centering
    \includegraphics[width=\linewidth]{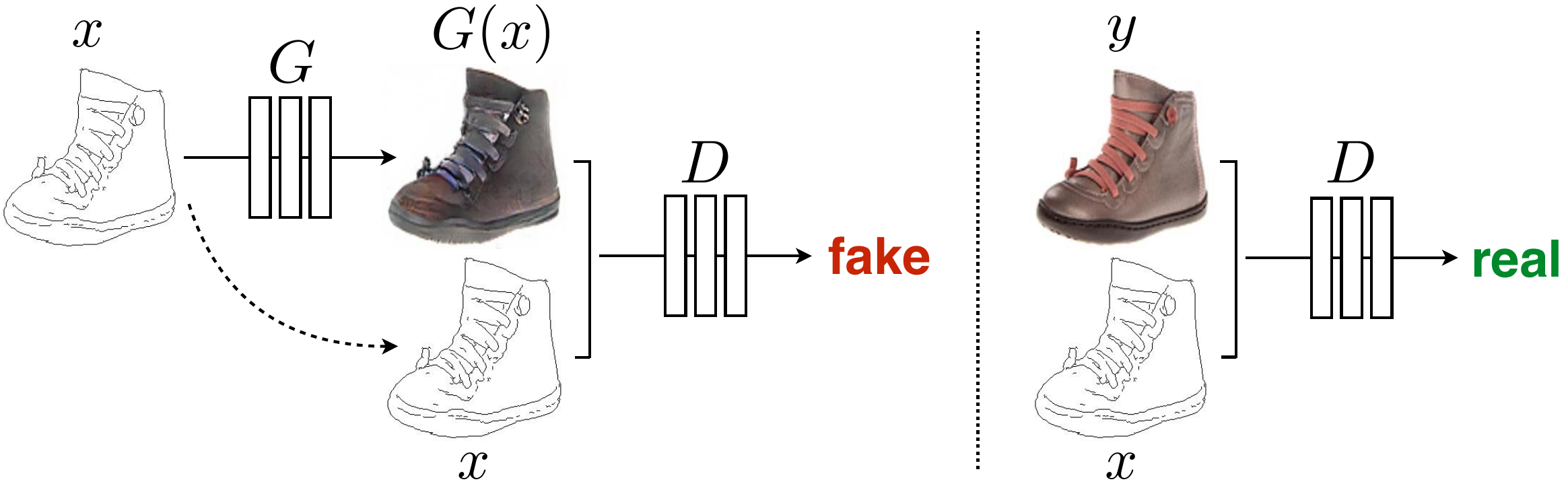}
    \caption{Pix2pix training procedure. Source-target domain pairs are used for training: the generator is an autoencoder that translates the input source domain to the target domain; the discriminator critic pairs of images composed of the source and (real or fake) target domains. Reproduced from Isola \emph{et al.} \cite{pix2pix}.}
    \label{fig:arch_pix2pix}
\end{figure}

Up until this time, the synthetic samples follow the same quality of plain generation's: low quality and low resolution.
This scenario changes with pix2pix \cite{pix2pix}. Pix2pix employed a new architecture for both the generator and the discriminator, as well as a new loss function. It was a complete revolution! We reproduce a simplification of the referenced architecture in Figure \ref{fig:arch_pix2pix}. The generator is a U-Net-like network \cite{ronneberger2015u}, where the skip connections allow to bypass information that is shared by the source-target pair. Also, the authors introduce a patch-based discriminator (which they called PatchGAN) to penalize structure at scale of patches of a smaller size (usually $70\times70$), while accelerating evaluation. To compose the new loss function, authors proposed the addition of a term that evaluates the L1 distance between synthetic and ground truth targets, constraining the synthetic samples without killing variability.

Despite advances on conditional plain generation techniques such as ACGAN \cite{odena2017conditional}, which allowed generation of samples up to $128\times128$ resolution, the quality of synthetic samples reached a new level with its contemporary pix2pix. This model is capable to generate $512\times512$ resolution synthetic images, comprising state-of-the-art levels of detail for that time. Overall, feeding the generator with the source sample's extra information simplify and guide generation, impacting the process positively.

\begin{figure*}[t]
    \centering
    \includegraphics[width=0.9\linewidth]{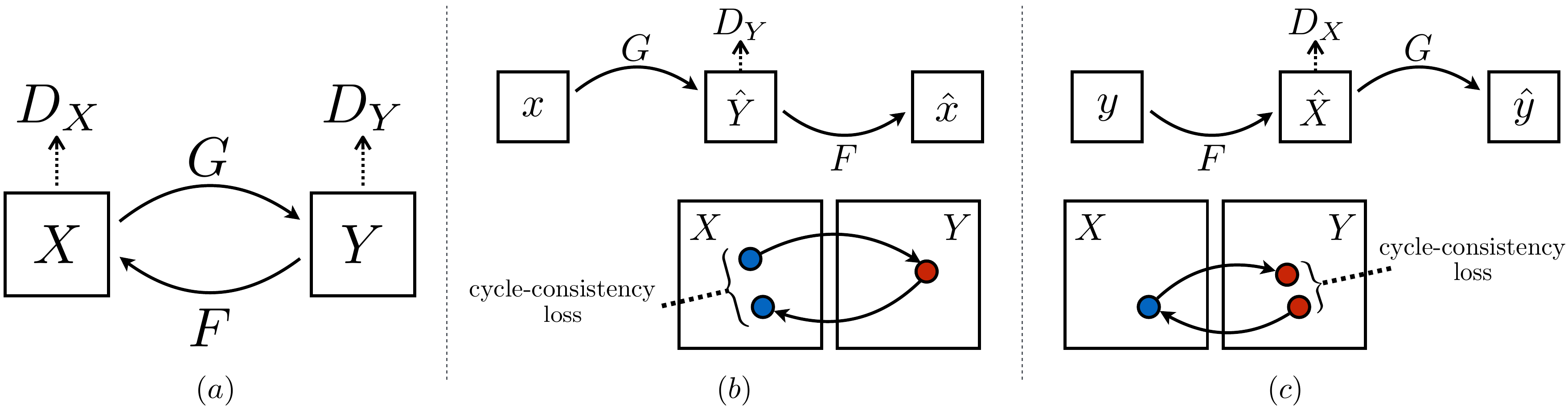}
    \caption{(a) CycleGAN uses two Generators ($G$ and $F$) and two discriminators ($D_X$ and $D_Y$) to learn to transform the domain $X$ into the domain $Y$ and vice-versa. To evaluate the cycle-consistency loss, authors force the generators to be able to reconstruct the image from the source domain after a transformation. That is, given domains $X$ and $Y$, generators $G$ and $F$ should be able to: (b) $X \rightarrow G(X) \rightarrow F(G(X)) \approx X$ and (c) $Y \rightarrow F(Y) \rightarrow G(F(Y)) \approx Y$.  Reproduced from Zhu \emph{et al.} \cite{zhu2017unpaired}.}
    \label{fig:arch_cyclegan}
\end{figure*}

The same research group responsible for pix2pix later released CycleGAN \cite{zhu2017unpaired}, further improving the overall quality of the synthetic samples. The new training procedure (see Figure \ref{fig:arch_cyclegan}) forces the generator to make sense of two translation processes: from source to target domain, and also from target to source. The cyclic training also uses separate discriminators to deal with each one of the translation processes. The architectures on the discriminators are the same from pix2pix, using $70\times70$ patches, while the generator receives the recent architecture proposed by Johnson \emph{et al.} \cite{johnson2016perceptual} for style transfer.

CycleGAN increased the domain count managed (generated) by a single GAN to two, making use of two discriminators (one for each domain) to enable translation between both learned domains. Besides the architecture growth needed, another limitation is the requirement of having pairs of data connecting both domains. Ideally, we want to increase the domain count without scaling the number of generators or discriminators proportionally, and have partially-labeled datasets (that is, not having pairs for every source-target domains). Those flaws motivated StarGAN \cite{choi2018stargan}. Apart from the source domain image, StarGAN's generator receives an extra array containing labels' codification that informs the target domain. This information is concatenated depth-wise to the source sample before feeding the generator, which proceeds to perform the same cyclic procedure from CycleGAN, making use of a reconstruction loss. To deal with multiple classes, without increasing the discriminator count, it accumulates a classification task to evaluate the domain of the analyzed samples.

The next step towards high-resolution image-to-image translation is pix2pixHD (High-Definition) \cite{pix2pixhd},
which obviously is based upon pix2pix's work but includes several modifications while adopting changes brought by CycleGAN with respect to the generator's architecture. 

\begin{figure}[t]
    \centering
    \includegraphics[width=\linewidth]{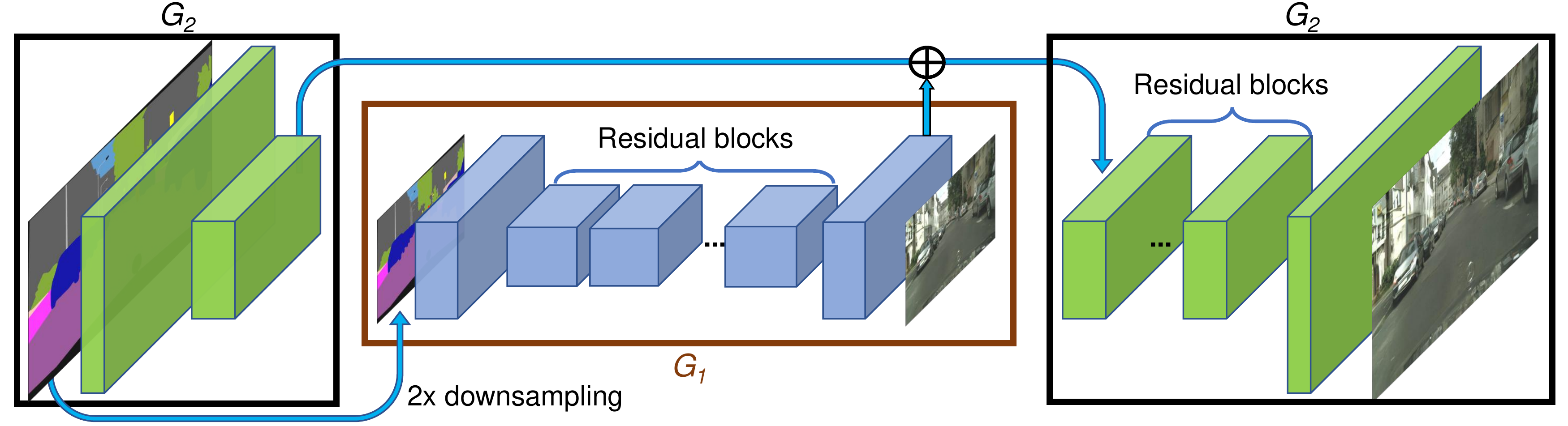}
    \caption{Summary of Pix2pixHD generators. $G_1$ is the global generator and $G_2$ is the local generator. Both generators are first trained separately, starting from the lower-resolution global generator, and next, proceeding to the local generator. Finally, both are fine-tuned together to generate images up to $2048\times1024$ resolution. Reproduced from Wang \emph{et al.} \cite{pix2pixhd}.}
    \label{fig:archpix2pixhd}
\end{figure}

The authors propose using two nested generators to enable the generation of $2048\times1024$ resolution images (see Figure \ref{fig:archpix2pixhd}), where the outer ``local'' generator enhances the generation of the inner ``global'' generator. Just like CycleGAN, it uses Johnson \emph{et al.} \cite{johnson2016perceptual} style transfer network as global generator, and as base for the local generator. The output of the global generator feeds the local generator in the encoding process (element-wise sum of global's features and local's encoding) to carry information of the lower resolution generation. They are also trained separately: first they train the global generator, then the local, and finally, they fine-tune the whole framework together.  

In pix2pixHD, the discriminator also receives upgrades. Instead of working with lower-resolution patches, pix2pixHD uses three discriminators that work simultaneously in different resolutions of the same images. This way, the lower resolution discriminator will concern more about the general structure and coarse details, while the high-resolution discriminators will pay attention to fine details.
The loss function also became more robust: besides the traditional adversarial component for each of the discriminators and generators, it comprises feature matching and perceptual loss components.

Other less structural aspects of generation, but maybe even more important, were explored by Wang \emph{et al.} \cite{pix2pixhd}. Usually, the input for image-to-image translation networks is semantic maps~\cite{lin2014microsoft}. 
It is an image where every pixel has the value of its object class and they are often the result of pixelwise segmentation tasks. During evaluation, the user can decide and pick the desired attributes of the result synthetic image by crafting the input semantic map. However, pix2pixHD authors noticed that sometimes this information is not enough to guide the generation. For example, let us think of the semantic map containing a queue of cars in the street. The blobs corresponding to each car would be connected, forming a strange format (for a car) blob, making it difficult for the network to make sense of it.

The authors' proposed solution is to add an instance map to the input of the networks. The instance map \cite{lin2014microsoft} is an image where the pixels combine information from its object class and its instance. Every instance of the same class receives a different pixel value. The addition of instance maps is one of the factors that affected the most our skin lesion generation, as well as other contexts showed in their~paper. 

A drawback of pix2pixHD and the other methods so far is that generation is deterministic (\emph{e.g.}, in test time, for a given source sample, the result is always the same). This is an undesired behavior if we plan to use the synthetic samples to augment a classification model's training data, for example.

Huang \emph{et al.} introduce Multimodal Unsupervised Image-to-image Translation (MUNIT) \cite{huang2018multimodal} to generate diverse samples with the same source sample. For each domain, the authors employ an encoder and a decoder to compose the generator. The main assumption is that it is possible to extract two types of information from samples: content, that is shared among instances of different domains, controlling general characteristics of the image; and style, that controls fine details that are specific and unique to each domain.
The encoder learns to extract content and style information, and the decoder to take advantage of this information. 

\begin{figure}[t]
    \centering
    \includegraphics[width=\linewidth]{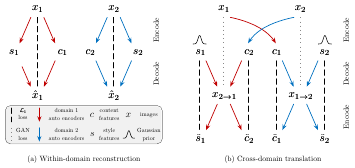}
    \caption{MUNIT's training procedure. (a) Reconstruction loss with respect to the images of the same domain. Style ``s'' and content ``c'' information are extracted from the real image, and used to generate a new one. The comparison of both images composes the model's loss function. (b) For cross-domain translation, the reconstruction of the latent vectors containing style and content information also compose the loss function. The content is extracted from a source real image, and the style is sampled from a Gaussian prior. Reproduced from Huang \emph{et al.} \cite{huang2018multimodal}.}
    \label{fig:arch_munit2}
\end{figure}

During training, two reconstruction losses are evaluated: the image reconstruction loss, which measure the ability to reconstruct the source image using the extracted content and style latent vectors; and the latent vectors reconstruction loss, which measure the ability to reconstruct the latent vectors themselves, comparing a pair of source latent vectors sampled from a random distribution, with the encoding of a synthetic image created using them (see Figure \ref{fig:arch_munit2}).

\begin{figure}[t]
    \centering
    \includegraphics[width=\linewidth]{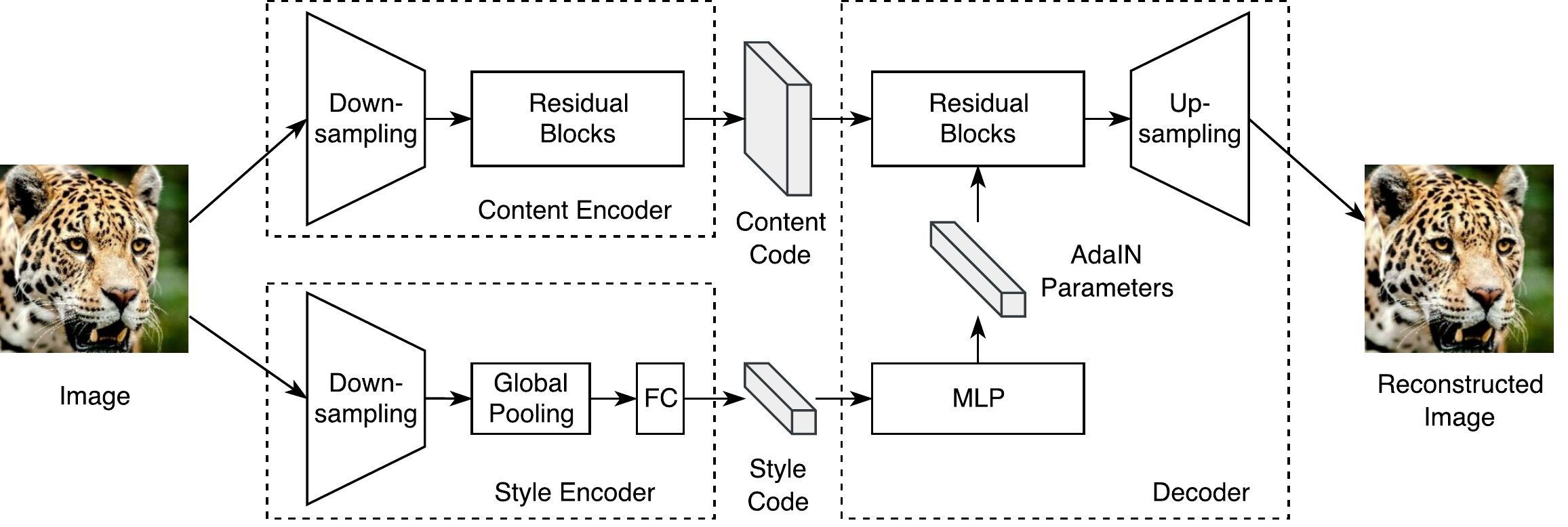}
    \caption{The generator's architecture from MUNIT. The authors employ different encoders for content and style. The style information feeds a Multi-Layer Perceptron (MLP) that provide the parameters for the AdaIN normalization. This process incorporates the style information to the content, creating a new image. Reproduced from Huang \emph{et al.} \cite{huang2018multimodal}.}
    \label{fig:arch_munit1}
\end{figure}

MUNIT's decoder (see Figure \ref{fig:arch_munit1}) incorporate style information using AdaIN \cite{huang2017arbitrary} (we detailed AdaIN, and the intuitions behind using this normalization in Section \ref{sec:sotaclass}), and it directly influenced current state-of-the-art GAN architectures \cite{karras2019style, park2019semantic, liu2019few}. 

One of the influenced works deals with a slightly different task: few-shot image-to-image translation. Few-shot translation attempts to translate a source image to a new unseen (but related) target domain after looking into just a few ($2$ or so) examples during test time (\emph{e.g.}, train with multiple dog breeds, and test for lions, tigers, cats, wolves).
To work on this problem, Liu \emph{et al.} \cite{liu2019few} introduce Few-shot Unsupervised Image-to-Image Translation (FUNIT). 
It combines and enhances methods from different GANs we already described. 
Authors propose an extension of CycleGAN's cyclic training procedure to multiple source classes (authors advocate that the higher the amount, the best the model's generalization); adoption of MUNIT's encoders for content and style, that are fused through AdaIN layers; enhancement of StarGAN's procedure to feed class information to the generator in addition to the content image, where instead of simple class information, the generator receives a set of few images of the target domain. The discriminator also follows StarGAN's, in a way that it performs an output for each of the source classes. This is an example of how works influence each other, and of how updating an older idea with enhanced recent techniques can result in a state-of-the-art solution.

So far, every image-to-image translation GAN generator's assumed the form of an autoencoder, where the source image is encoded into a reduced latent representation, that is finally expanded to its full resolution. The encoder plays an important role to extract information of the source image that will be kept in the output. Often, even multiple encoders are employed to extract different information, such as content and style. 

\begin{figure}[t]
    \centering
    \includegraphics[width=0.6\linewidth]{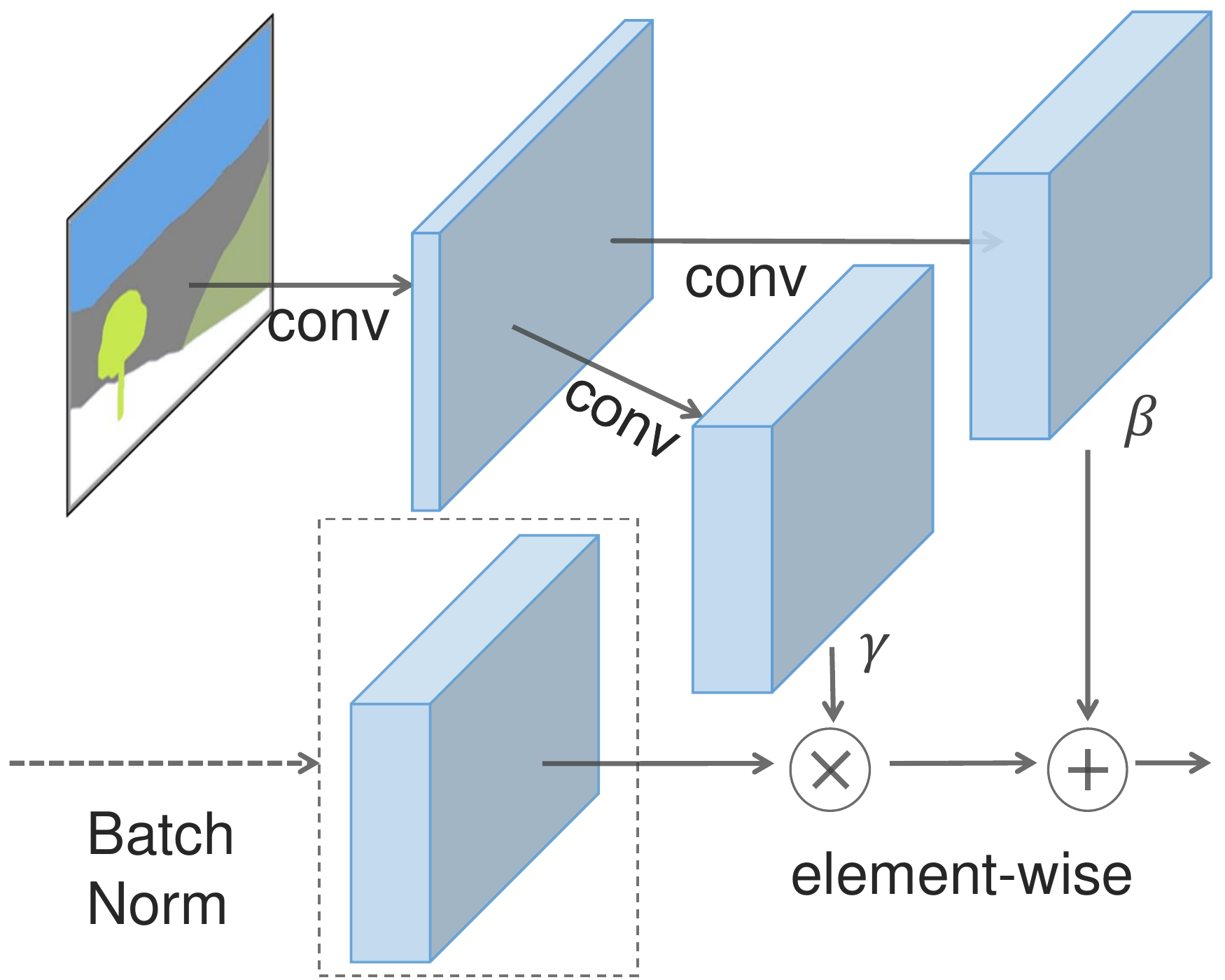}
    \caption{At the SPADE normalization process, the input semantic map is first projected into a feature space. Next, different convolutions are responsible for extracting the parameters that perform element-wise multiplication and sum to the normalized generator's activation. Reproduced from Park \emph{et al.} \cite{park2019semantic}.}
    \label{fig:spadenorm}
\end{figure}

\begin{figure}[t]
    \centering
    \includegraphics[width=0.65\linewidth]{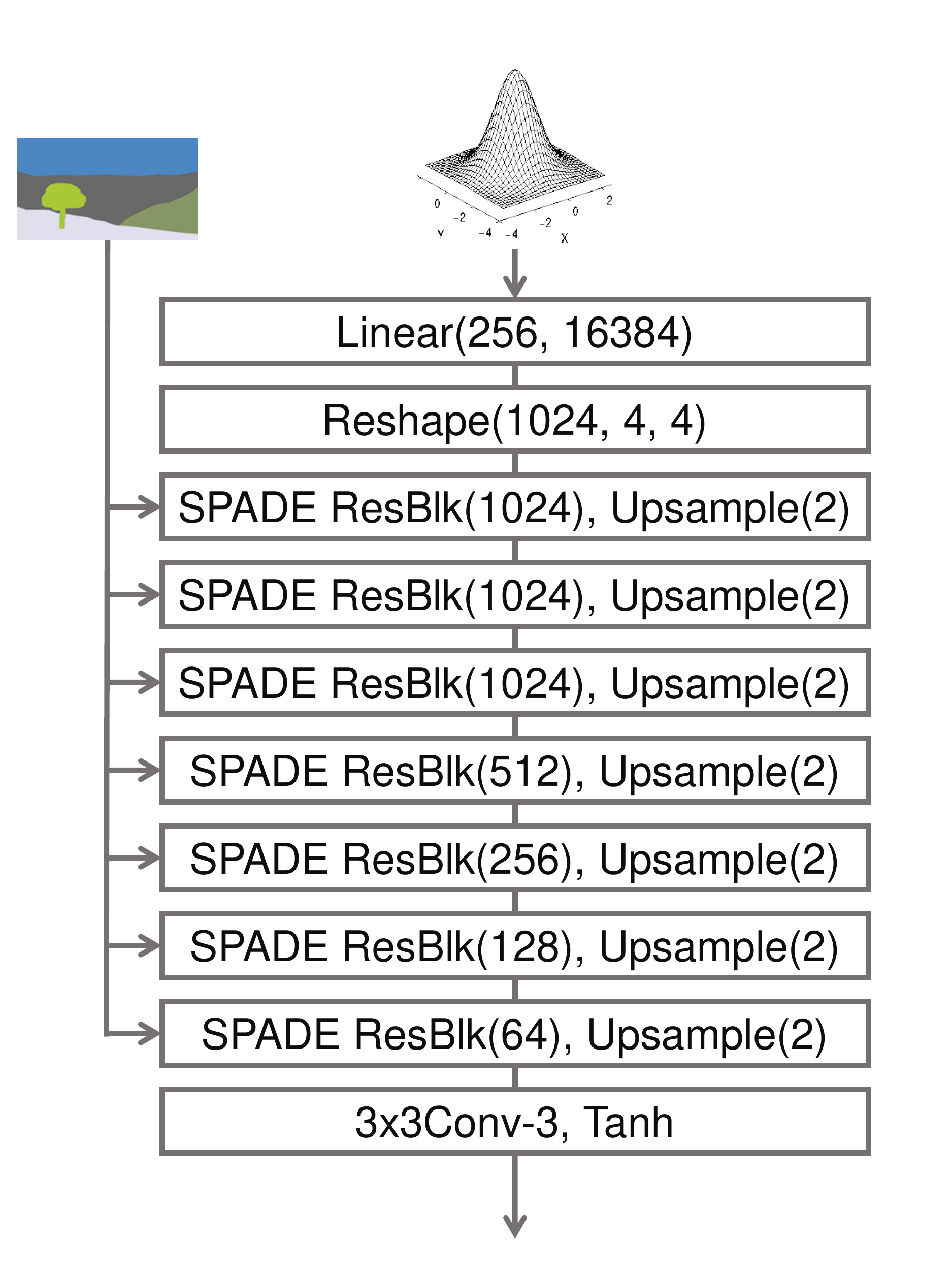}
    \caption{SPADE's generator. A sampled noise is modified through the residual blocks. After each block, the output is shifted and scaled using AdaIN. This way, the style information of the input semantic map is present in every stage of the generation, without killing the output variability that comes with the sampled noise at the beginning of the process. Reproduced from Park \emph{et al.} \cite{park2019semantic}.}
    \label{fig:spadegen}
\end{figure}

On Spatially-Adaptive Normalization (SPADE) \cite{park2019semantic} authors introduce a method for semantic image synthesis (\emph{e.g.}, image-to-image translation using a semantic map as the generator's input). It can be considered pix2pixHD \cite{pix2pixhd} successor, in a way that it deals with much of the previous work flaws. Although the authors call their GAN as SPADE in the paper (they call it GauGAN now), the name refers to the introduced normalization process, which generalizes other normalization techniques (\emph{e.g.}, Batch Normalization, Instance Normalization, AdaIN). Like AdaIN, SPADE is used to incorporate the input information into the generation process, but there is a key difference between both methods. On SPADE, the parameters used to shift and scale the feature maps are tensors that contain spatial information preserved from the input semantic map. Those parameters are obtained through convolutions, then multiplied and summed element-wise to the feature map (see Figure \ref{fig:spadenorm}). This process takes place in all the generator's layers, except for the last one, which outputs the synthetic image. Since the input of the generator's decoder is not the encoding of the semantic map, authors use noise to feed the first generator's layer (see Figure \ref{fig:spadegen}). This change enables SPADE for multimodal generation, that is, given a target semantic map, SPADE can generate multiple different samples using the same map.

\subsection{Validation Metrics of Generative Methods} \label{sec:sotavalidation}

\noindent \textit{Inception Score} (IS) \cite{salimans2016}: it uses an Inceptionv3 network pre-trained on ImageNet to compute the logits of the synthetic samples. The logits are used to evaluate Equation~\ref{eq:is}. The authors say it correlates well with human judgment over synthetic images.
Since the network was pre-trained on ImageNet, we rely on its judgment of the synthetic images with respect to the ImageNet classes. This is a big problem because skin lesion images do not relate to any class on ImageNet. Thus, this method is inappropriate to evaluate synthetic samples from any dataset that is not ImageNet.

\begin{equation}
    \begin{aligned}
          \is (G) = \exp \left(\mean_{\mathbf{x}\backsim p_g} [D_{\text{KL}}(p(y|\mathbf{x}) \parallel p(y))] \right), \\
        p(y) = \int_\mathbf{x} p(y|\mathbf{x})p_g(\mathbf{x}),
    \label{eq:is}
    \end{aligned}
\end{equation}
\noindent where $\mathbf{x} \sim p_g$ indicates that $\mathbf{x}$ is an image sampled from $p_g$, $p_g$ is the distribution learned by the generator, $D_{KL}(p \| q)$ is the Kullback Leibler divergence between the distributions $p$ and $q$, $p(y|\mathbf{x})$ is the conditional class distribution (logit of a given sample), and $p(y)$ is the marginal class distribution (mean logits over all synthetic samples).

\noindent \textit{Frech\`et Inception Distance} (FID) \cite{heusel2017gans}: like the Inception Score, the FID relies on Inception's evaluation to measure quality of synthetic samples and suffer from the same problems. Differently though, it takes the features from the Inception's penultimate layer from both real and synthetic samples, comparing them. The FID uses Gaussian approximations for these distributions, which makes it less sensitive to small details (which are abundant in high-resolution samples). 

\noindent \textit{Sliced Wasserstein Distance} (SWD) \cite{karras2017progressive}: Karras \emph{et al.} introduced the SWD metric to deal specifically with high-resolution samples. The idea is to consider multiple resolutions for each image, going from $16\times16$ and doubling until maximum resolution (Laplacian Pyramid). For each resolution, slice $128$ $7\times7\times3$ patches from each level, for both real and synthetic samples. Finally, use the Sliced Wasserstein Distance \cite{rabin2011wasserstein} to evaluate an approximation to the Earth Mover's distance between both. 

\noindent \textit{GANtrain and GANtest} \cite{shmelkov2018good}: the idea behind these metrics align with our objective of using synthetic images as part of a classification network, like a smarter data augmentation process. GANtrain is the accuracy of a classification network trained on the synthetic samples, and evaluated on real images. Similarly, GANtest is the accuracy of a classification network trained on real data, and evaluated on synthetic samples. The authors compare the performance of GANtrain and GANtest with a baseline network trained and tested on real data.

Borji \cite{borji2019pros} analyzed the existing metrics in different criteria: discriminability (capacity of favoring high-fidelity images), detecting overfitting, disentangled latent spaces, well-defined bounds, human perceptual judgments, sensitivity to distortions, complexity and sample efficiency. After an extensive review of the metrics literature, the author compares the metrics concerning the presented criteria, and among differences and similarities, can not point the definitive metric to be used. The author suggests future studies to rely on different metrics to better assess the quality of the synthetic images. 

Theis \emph{et al.} \cite{theis2015note}, in a study of quality assessment for synthetic samples, highlighted that the same model may have very different performance on different applications, thus, a proper assessment of the synthetic samples must consider the context of the application.

\section{Conclusions}
Despite the advancements achieved in the last years, enabling the generation of human faces that are indistinguishable from real people, there are many problems to solve. Mode collapse is still present in all proposed GANs, and the problem becomes even more critical when the number of classes is high, or for unbalanced datasets. The advancements in this problem can be decisive for GANs to be more employed in real scenarios, leaving the academic environment. 

In this sense, there is another major innovation that should be achieved in the next few years: exploring techniques to make better use of the parameters, making GANs lighter, and feasible to run in mobile devices without losing (much) performance. To show how far we are from this reality today, on GauGAN \cite{park2019semantic} only the generator can get to more than 120 million parameters. Parallel to this reality, convolutional neural networks are already moving towards this direction. Bottleneck modules, for example, aid in decreasing the number of parameters (although they used it as an excuse to stack bottleneck modules, increasing the number of parameters again) and are present in most modern CNN architectures. Also, entire networks for classification and segmentation can today be trained on mobile devices, enabling these solutions to be used in a much bigger set of problems.

\section*{Acknowledgment}
A. Bissoto and S. Avila are partially funded Google Latin America Research Awards (LARA) 2018. A. Bissoto was also partially funded by CNPq (134271/2017-3). S. Avila is also partially funded by FAPESP (2017/16246-0). E. Valle is partially funded by CNPq grants (PQ-2 311905/2017-0, 424958/2016-3) and FAPESP (2019/05018-1). The RECOD Lab receives addition funds from FAPESP, CNPq, and CAPES. We gratefully acknowledge NVIDIA for the donation of GPU hardware.
\bibliographystyle{unsrt}
\bibliography{mainbib}

\end{document}